\def\year{2021}\relax
\documentclass[letterpaper]{article} 
\usepackage{aaai21}  
\usepackage{times}  
\usepackage{helvet} 
\usepackage{courier}  
\usepackage[hyphens]{url}  
\usepackage{graphicx} 
\usepackage{natbib}  
\usepackage{caption} 
\usepackage{rotating,tabularx}
\usepackage{multirow}
\usepackage{float}
\usepackage{booktabs} 
\usepackage{url}
\usepackage{graphicx}  
\usepackage{amsfonts} 
\usepackage{amsmath}
\usepackage{bm}
\usepackage{multirow}
\usepackage{subfigure}
\usepackage{algorithm}  
\usepackage{algpseudocode}  
\usepackage{makecell}
\usepackage{xspace}
\usepackage{amssymb}
\usepackage{array}
\usepackage{footmisc}
\usepackage{listings}
\usepackage[np, autolanguage]{numprint}
\usepackage{paralist}
\usepackage{enumitem}
\usepackage{colortbl}
\usepackage{soul}
\usepackage{epstopdf}
\usepackage{pifont}
\usepackage{xcolor}
\usepackage{mathrsfs}
\usepackage{todonotes}
\usepackage{subfig}
\usepackage{CJK}

\newcommand{\ie}{\emph{i.e.,}\xspace}

\newcommand{\cbkgrnd}{\cellcolor{blue!15}}

\newcommand{\dubbelop}{$^{\blacktriangle}$}

\newcommand{\dubbelneer}{$^{\blacktriangledown}$}

\newcommand{\hlc}[2][yellow]{{%
    \colorlet{foo}{#1}%
    \sethlcolor{foo}\hl{#2}}%
}

\usepackage{microtype}

\makeatletter 
\newcommand\figcaption{\def\@captype{figure}\caption} 
\newcommand\tabcaption{\def\@captype{table}\caption}

\makeatletter
\newcommand{\thickhline}{%
	\noalign {\ifnum 0=`}\fi \hrule height 1pt
	\futurelet \reserved@a \@xhline
}
\makeatother

\title{The Style-Content Duality of Attractiveness:\\ Learning to Write Eye-Catching Headlines via Disentanglement}

\author {

        Mingzhe Li \textsuperscript{\rm 1, \rm 2,}\thanks{Equal contribution.},
        Xiuying Chen \textsuperscript{\rm 1, \rm 2,}\footnotemark[1],
        Min Yang \textsuperscript{\rm 3},
        Shen Gao \textsuperscript{\rm 2},
        Dongyan Zhao \textsuperscript{\rm 1, \rm 2},
        Rui Yan \textsuperscript{\rm 4,5,}\thanks{Corresponding author: Rui Yan (ruiyan@ruc.edu.cn)}\\
}
\affiliations {
    \textsuperscript{\rm 1} Center for Data Science, AAIS, Peking University,Beijing,China \\
    \textsuperscript{\rm 2} Wangxuan Institute of Computer Technology, Peking University,Beijing,China \\
    \textsuperscript{\rm 3} Shenzhen Institutes of Advanced Technology, Chinese Academy of Sciences, Shenzhen, China \\
    \textsuperscript{\rm 4} Gaoling School of Artificial Intelligence, Renmin University of China \\
    \textsuperscript{\rm 5} Beijing Academy of Artificial Intelligence \\
    \{li\_mingzhe,xy-chen,shengao,zhaody\}@pku.edu.cn,
    min.yang@siat.ac.cn, ruiyan@ruc.edu.cn

}
\begin{document}
  \maketitle
  \begin{abstract}
Eye-catching headlines function as the first device to trigger more clicks, bringing reciprocal effect between producers and viewers.
Producers can obtain more traffic and profits, and readers can have access to outstanding articles.
When generating attractive headlines, it is important to not only capture the attractive \textit{content} but also follow an eye-catching written \textit{style}. 
In this paper, we propose a Disentanglement-based Attractive Headline Generator (DAHG) that generates headline which captures the attractive content following the attractive style.
Concretely, we first devise a disentanglement module to divide the style and content of an attractive prototype headline into latent spaces, with two auxiliary constraints to ensure the two spaces are indeed disentangled.
The latent content information is then used to further polish the document representation and help capture the salient part.
Finally, the generator takes the polished document as input to generate headline under the guidance of the attractive style. 
Extensive experiments on the public Kuaibao dataset show that DAHG achieves state-of-the-art performance. 
Human evaluation also demonstrates that DAHG triggers 22\% more clicks than existing models.
  \end{abstract}
  
  \section{Introduction}
With the rapid growth of information spreading throughout the Internet, readers get drown in the sea of documents, and will only pay attention to those articles with attractive headlines that can catch their eyes at first sight.
On one hand, generating headlines that can trigger high click-rate is especially important for different avenues and forms of media to compete for user's limited attention.
On the other hand, only with the help of a good headline, can the outstanding article be discovered by readers.

To generate better headlines, we first analyze what makes the headlines attractive.
By surveying hundreds of headlines of popular websites, we found that one important feature that influences the attractiveness of a headline is its \textbf{\textit{content}}.
For example, when reporting the same event, the headline ``Happy but not knowing danger: Children in India play on the poisonous foam beach'' wins over 1000 page views, while the headline ``Chennai beach was covered with white foam for four days in India'' only has 387 readers.
The popular headline highlights the fact that ``the beach is poisonous and affects children'',  which will concern more people than ``white foam''.
On the other hand, the \textbf{\textit{style}} of the headline also has a huge impact on attractiveness.
For example, the headline ``Only two people scored thousand in the history of NBA Finals'' attracts fewer people than the headline ``How hard is it to get 1000 points in the NBA finals? Only two people in history!'', due to its conversational style that makes readers feel the need to see the answer to this question.
    
Most of the recent researches regard the headline generation task merely as a typical summarization task \cite{shu2018deep}.
This is not sufficient because a good headline should not only capture the most relevant content of an article but also be attractive to the reader.
However, attractive headline generation tasks were paid less attention by researchers.
\citet{xu2019clickbait} tackle this task by adversarial training, using an attractiveness score module to guide the summarization process.
\citet{jin2020hooks} introduce a parameter sharing scheme to disentangle the attractive style from the attractive text.
However, previous works neglect the fact that attractiveness is not just about \textit{style}, but also about \textit{content}.

Based on the above analysis, we propose a model named \emph{Disentanglement-based Attractive Headline Generation} (DAHG), which learns to write attractive headlines from both style and content perspectives. 
These two attractiveness attributes are learned from an attractive prototype headline, \ie the headline of the document in the training dataset that is most similar to the input document.
First, DAHG separates the attractive style and content of the prototype headline into latent spaces, with two auxiliary constraints to ensure the two spaces are indeed disentangled.
Second, the learned attractive content space is utilized to iteratively polish the input document, emphasizing the parts in the document that are attractive.
Finally, the decoder generates an attractive headline from the polished input document representation under the guidance of the separated attractive style space.
Extensive experiments on the public Kuaibao dataset show that DAHG outperforms the summarization and headline generation baselines in terms of ROUGE metrics, BLEU metrics, and human evaluations by a large margin.
Specifically, DAHG triggers 22\% more clicks than the strongest baseline.

The major contributions of this paper are as follows: 
(1) We devise a disentanglement mechanism to divide the attractive content and style space from the attractive prototype headline.
(2) We propose to generate an attractive headline with the help of disentangled content space under the style guidance.
(3) Experimental results demonstrate that our model outperforms other baselines in terms of both automatic and human evaluations.
  
  \section{Related Work}
  Our research builds on previous works in three fields: text summarization, headline generation, and disentanglement.
  
  \begin{figure*}
    \centering
    \includegraphics[scale=0.6]{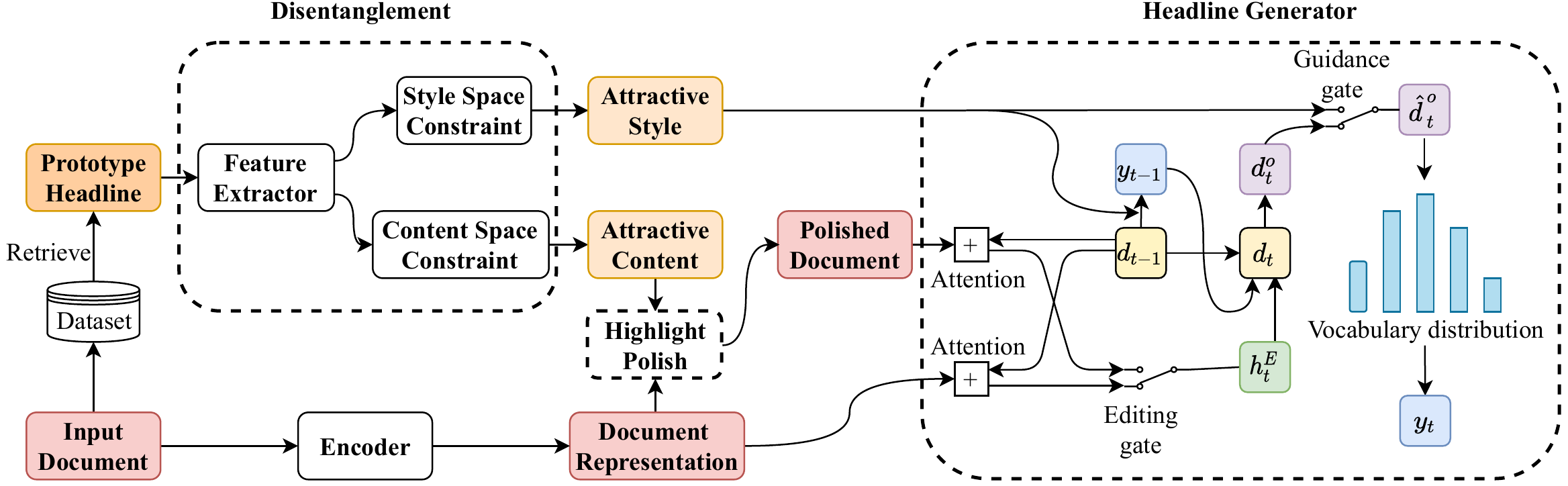}
    \caption{
      Overview of DAHG. We divide our model into three parts: (1) \textit{Disentanglement} disentangles attractive style and attractive content space from prototype headline representation; (2)
      \textit{Highlight Polish} highlights the attractive part in the input document with the help of attractive content space; (3) \textit{Headline Generator} generates headline taking the polished document as input under the guidance of attractive style.
    }
    \label{fig:overview}
  \end{figure*}
    
    \noindent \textbf{Text Summarization.}
    Headline generation is a task based on text summarization, where methods can be divided into two categories, extractive, and abstractive methods. 
    Extractive models~\cite{nallapati2017summarunner, zhou2018neural, zhang2018neural} directly select sentences from article as the summary.
    In contrast, abstractive models ~\cite{ gao2019abstractive,chen2019learning,Gao2020From,gao2020standard,li2020vmsmo} generate a summary from scratch.
    A series of work relies on prototype text to assist summarization.
    \citet{cao2018retrieve} chose the template most similar to the input as a soft template to generate summaries.
    Following this, \citet{gao2019write} proposed to generate the summary with the pattern based on prototype editing.
    Our work differs from previous works in focusing on the attractiveness of the generated summary.
    
    \noindent \textbf{Headline Generation.}
    In recent years, text generation has made impressive progress~\cite{li2019learning,chan2019stick,liu2020character,xie2020infusing,chan2020selection,chen2021reasoning}, and headline generation has become a research hotspot in Natural Language Processing.
    Most existing headline generation works solely focus on summarizing the document.
    \citet{tan2017neural} exploited generating headlines with hierarchical attention.
    \citet{gavrilov2019self} applied recent universal Transformer~\cite{dehghani2018universal} architecture for headline generation.
    Attractive headline generation was paid less attention by researchers.
    \citet{xu2019clickbait} trained a sensation scorer to judge whether a headline is attractive and then used the scorer to guide the headline generation by reinforcement learning.
    \citet{jin2020hooks} introduced a parameter sharing scheme to further extract attractive style from the text.
    However, to our best knowledge, no existing work considers the style-content duality of attractiveness.

    \noindent \textbf{Disentanglement.}
Disentangling neural networks' latent space has been explored in the computer vision domain, and researchers have successfully disentangled the features (such as rotation and color) of images \cite{chen2016infogan,higgins2017beta}. 
    Compared to the computer vision field, NLP tasks mainly focus on invariant representation learning.
    Disentangled representation learning is widely adopted in non-parallel text style transfer. 
    For example, \citet{fu2018style} proposed an approach to train style-specific embedding. 
    Some work also focused on disentangling syntax and semantic representations in text.
    \citet{iyyer2018adversarial} proposed syntactically controlled paraphrase networks to produce a paraphrase of the sentence with the desired syntax given a sentence and a target syntactic form.
    \citet{chen2019multi} proposed a generative model to get better syntax and semantics representations by training with multiple losses that exploit aligned paraphrastic sentences and word-order information.
    
    Existing works concentrate on learning the disentangled representation, and we take one step further to utilize this representation to generate attractive headlines.

  \section{Problem Formulation}
Before presenting our approach for the attractive headline generation, we first introduce our notations and key concepts. 
For an input document $X^d=\{x_1^d, x_2^d, \dots , x_{m^d}^d\}$ which has ${m^d}$ words, we assume there is a corresponding headline $Y^d=\{y_1^d, y_2^d, \dots, y_{n^d}^d\}$ which has ${n^d}$ words.
In our setting, the most similar document-headline pair to the current document-headline pair is retrieved from the training corpus as prototype pair.
Retrieve details are introduced in \S\ref{dataset}.
The prototype document is defined as $X^r=\{x_1^r, x_2^r, \dots, x_{m^r}^r\}$, and prototype headline is $Y^r=\{y_1^r, y_2^r, \dots, y_{n^r}^r\}$.
  
For a given document $X^d$, our model first divides the latent representation of the retrieval attractive prototype headline $Y^r$ into two parts: the attractive style space $s$, and attractive content space $c$.
Then we use the content $c$ to extract and polish document $X^d$ and use the style $s$ to guide attractive headline generation.
The goal is to generate a headline $\hat{Y}^d$ that not only covers the salient part of input document $X^d$ but also follows an attractive style.
  
\section{Model}
\subsection{Overview}
In this section, we propose our Disentanglement-based Attractive Headline Generation (DAHG) model, which can be divided into three parts as shown in Figure~\ref{fig:overview}:

$\bullet$ \textbf{Disentanglement} contains (1) \textit{Feature Extractor} module, which projects the prototype headline representation into latent style and content space,
(2) \textit{Style Space Constraint}, and (3) \textit{Content Space Constraint} module, which ensure that the extracted style and content space does not mix with each other.
    
$\bullet$ \textbf{Highlight Polish} highlights the attractive part of the document guided by the attractive content space learned from the prototype headline. 
    
$\bullet$ \textbf{Headline Generator} generates an attractive headline based on the polished document under the guidance of the attractive style.

\subsection{Disentanglement}
\begin{figure}
  \centering
  \includegraphics[scale=0.64]{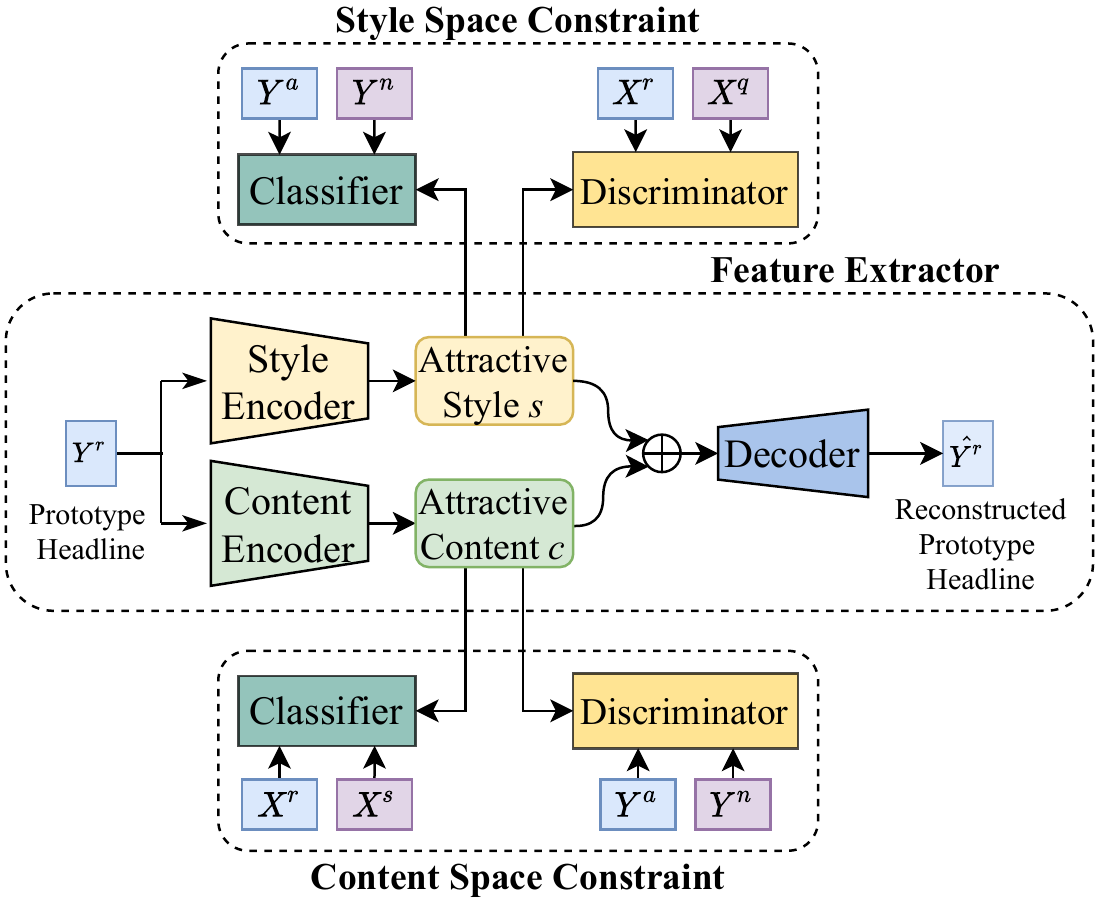}
  \caption{
    Framework of disentanglement module. 
$Y^r$ is the input prototype headline and $\hat{Y}^r$ is the reconstructed prototype headline.
$X^{r}$ denotes the prototype document, $X^{s}$ denotes the most similar document with $X^{r}$, and $X^{q}$ denotes a random document.
$Y^a$ is a random attractive headline and $Y^n$ is a random unattractive headline.
  }
  \label{fig:disentangle}
\end{figure}

The disentanglement framework shown in Figure~\ref{fig:disentangle} consists of three components.
The feature extractor serves as an autoencoder, with two constraints to facilitate disentanglement.

To begin with, we map a one-hot representation of each word in input document $X^d$ and prototype document $X^r$ into a high-dimensional vector space and employ a bi-directional recurrent neural network ($\rm{\text{Bi-RNN}_X}$) to model the temporal interactions between words. 
Then the encoder states of $X^d$ and $X^r$ is represented as $h_t^{X^d}$ and $h_t^{X^r}$, respectively.
Following~\citet{see2017get}, we choose Long Short-Term Memory (LSTM) \cite{hochreiter1997long} as the cell for Bi-RNN. 
We concatenate the last hidden state of the forward RNN $h^{X^r}_{m^r}$ and backward RNN $h^{X^r}_1$ and obtain the overall representation vector $h^{X^r}$  for $X^r$.

\textbf{Feature Extractor.}
We employ Variational AutoEncoder (VAE) as feature extractor, since VAE is appealing in explicitly modeling global properties such as syntactic, semantic, and discourse coherence \cite{li2015hierarchical, yu2020draft}.
Concretely, feature extractor consists of two encoders and a decoder.
The two feature encoders map the input prototype headline $Y^r$ to an attractive content space $c$ and an attractive style space $s$, and the decoder reconstructs the prototype headline $\hat{Y}^r$ from $c$ and $s$.
  
Concretely, the two encoders compute two posterior distributions $q_\theta(c|Y^r)$ and $q_\theta(s|Y^r)$ given the prototype headline $Y^r$, corresponding to content space $c$ and style space $s$, respectively.
The latent representations $c$ and $s$ are obtained by sampling from $q_\theta(c|Y^r)$ and $q_\theta(s|Y^r)$.
The reconstruction process can be formulated as $p_\theta(Y^r|[c;s])$, representing the probability of generating input $Y^r$ conditioned on the combination of content $c$ and style $s$. 
Herein $\theta$ represents the parameters of the above encoders and reconstruction decoder.
Because of the intractable integral of the marginal likelihood $p_\theta(Y^r)$~\cite{kingma2013auto}, the posterior $q_\theta(c|Y^r)$ and $q_\theta(s|Y^r)$ are simulated by variational approximation $q_\phi(c|Y^r)$ and $q_\phi(s|Y^r)$, where $\phi$ is the parameters for $q$.
  
When learning the VAE, the objective is to maximize the variational lower bound of $\log p_\theta(Y^r)$:
\begin{align}
\small
\mathcal{L}_{VAE}
&={\lambda}_{KL_c}{\rm KL}(q_\phi(c|Y^r)\|p_\theta(c)) \notag\\
&+{\lambda}_{KL_s}{\rm KL}(q_\phi(s|Y^r)\|p_\theta(s)) \notag\\
&-{\rm E}_{q_\phi(c|Y^r),q_\phi(s|Y^r)}[{\rm log}p_
\theta(Y^r|[c;s])],
\end{align}
where the $\text{KL}(\cdot)$ denotes KL-divergence, the regularization for encouraging the approximated posterior $q_\phi(c|Y^r)$ and $q_\phi(s|Y^r)$ to be close to the prior $p_\theta(c)$ and prior $p_\theta(s)$, \ie standard Gaussian distribution.
${\rm E}[\cdot]$ is the reconstruction loss conditioned on the approximation posterior $q_\phi(c|Y^r)$ and $q_\phi(s|Y^r)$.

\textbf{Style Space Constraint.}
Overall, to ensure that the style information is stored in the style space $s$, while the content information is filtered, we first design a style space constraint applied on the VAE in feature extractor. Generally, we use a classifier to determine the style label of $s$, and a discriminator to identity which document $s$ corresponds to.
If the style space $s$ can be successfully classified as attractive headlines, and the corresponding document cannot be distinguished, then we can safely draw the conclusion that $s$ only contains style information.

To disentangle the style information, we first randomly select an attractive headline $Y^a$ and an unattractive headline $Y^n$ as the two candidates of the classifier.
Then we use the same embedding matrix to map each word into a high-dimensional vector space, and use $\rm{Bi\text{-}RNN_Y}$ to obtain the representation $h^{Y^a}$ and $h^{Y^n}$ for the title $Y^a$ and $Y^n$, respectively.
The last states of $\rm{\text{Bi-RNN}_Y}$ are concatenated as the overall representation vector $h^{Y^a}$ and $h^{Y^n}$.
In this way, the classification result is obtained as:
\begin{align}
    C_s(Y^{*})={\rm softmax}(W_{ss}[s;h^{Y^*}]+b_{ss}).
\end{align}
Since the classifier aims to maximize the probability of matching the randomly selected title $Y^a$ with attractive style, the loss of the classifier network is defined as:
\begin{align}
    \mathcal{L}_{Cs} &= -\log(C_s(Y^{a})) - \log(1-C_s(Y^{n})).
\end{align}

As for filtering the content information from the style space $s$, we achieve this goal following the generative adversarial way \cite{goodfellow2014generative}. Generally, we employ a discriminator to distinguish the document that corresponds to the prototype headline from two candidates, while the feature extractor is trained to encode the style space from which its adversary cannot predict the corresponding document.
The positive sample is the prototype document $X^r$, and the negative sample is a random document, denoted by  $X^{q}$.
Similar to before, we use the same embedding matrix and $\rm{Bi\text{-}RNN_X}$ to obtain the negative sample representation $h^{X^{q}}$.
We build a two-way softmax discriminator on the style space $s$ to predict its corresponding document as:
\begin{gather}
    D(X^{*})={\rm softmax}(W_{sc}[s;h^{X^{*}}]+b_{sc}),
    \label{softmax_dis}
\end{gather}
where $h^{X^{*}}$ can be the ground truth prototype document representation $h^{X^r}$ or the negative document representation $h^{X^{q}}$.
  
The training objective of the discriminator is to maximize the log-likelihood for correct classification, while the feature extractor aims to encode the style space where classifier cannot predict the correct label:
\begin{align}
    \mathcal{L}_S^d &=  - \log(D(X^{r})) - \log(1-D(X^{q})),\\
    \mathcal{L}_S^g &=  -\log(1-D(X^{r})).
\end{align}

In this way, the classifier ensures that the style space $s$ stores the attractive style information and the discriminator guarantees that it does not contain content information.

\textbf{Content Space Constraint.}
Similar to style space constraint, the purpose of content space constraint is to ensure that the content information of the prototype headline is stored in content space $c$ and style information is not.
Therefore, a classifier is utilized to predict which document the content space $c$ matches, and a discriminator is employed to distinguish its style label from attractive or unattractive.

For classifier, the prototype document $X^r$ and its most similar document $X^{s}$ are provided as the two candidates, which improves the difficulty of the classifier.
If the classifier can successfully identify the corresponding sample from extremely similar candidates, then we can say that they achieve good performance, and the content is successfully disentangled. The content-oriented loss to regularize the content space is obtained as:
\begin{align}
    C_c(X^{*})&={\rm softmax}(W_{cc}[c;h^{X^{*}}]+b_{cc}),\\
    \mathcal{L}_{Cc} &= -\log(C_c(X^{r})) - \log(1-C_c(X^{s})),
\end{align}
where $h^{X^{*}}$ can be the representation of $X^r$ and $X^{s}$ obtained by the same embedding martrix and $\rm{Bi\text{-}RNN_X}$.

For discriminator, positive and negative samples are the random attractive title $Y^a$ and unattractive title $Y^n$, respectively, which are selected in the same way as in style space constraint.
The training objective of the feature extractor is to encode the content space from which its adversary cannot predict the corresponding style label as shown in Equation~\ref{lcd}, while the discriminator is trained to maximize the log-likelihood for correct style label as Equation~\ref{lcg}:
\begin{align}
    D(Y^{*}) &= {\rm softmax}(W_{cs}[c;h^{Y^{*}}]+b_{cs}),\\
    \mathcal{L}_C^d &=  - \log(D(Y^{a})) - \log(1-D(Y^{n})),\label{lcd}\\
    \mathcal{L}_C^g &=  -\log(1-D(Y^{a})) \label{lcg}.
\end{align}


  \subsection{Highlight Polish}
  \label{sec:polish}
  Intuitively, the prototype attractive content space can help find attractive highlights in the document.
Hence, following~\citet{chen2018iterative}, we design a highlight polish module, utilizing the disentangled attractive content representation $c$ to polish the document representation.

  Polish module consists of an RNN layer made up of Selective Recurrent Units (SRUs), a modified version of the original GRU.
  Generally speaking, SRU decides which part of the information should be updated based on both the polished document and the attractive content representation:
  \begin{gather}
  h^p_t={\rm SRU}(h^{X^d}_t,c),
  \end{gather}
  where $h_t^p$ denotes the $t$-th hidden state, and $c$ is the attractive content space extracted from disentanglement module.
  Details of SRU can be found in Appendix~\ref{sec:sru}.

  \subsection{Headline Generator}
  The headline generator targets at generating a headline based on the polished document representation $h_t^{p}$ following the extracted style space $s$. 
  The final state of the input document representation $h^{X^d}$ is employed as the initial state $d_0$ of the RNN decoder, and the procedure of $t$-th generation is calculated as:
  \begin{gather}
  d_t={\rm \text{LSTM}_{dec}}(d_{t-1}, [e(y_{t-1}); h^{E}_{t-1}]).
  \end{gather}
  $d_t$ is the hidden state of the $t$-th decoding step, $e(y_{t-1})$ is the embedding of last generated word $y_{t-1}$, and $h^{E}_{t-1}$ is the context vector calculated by the standard attention
  mechanism~\cite{bahdanau2014neural} in Equation~\ref{contextvector}.
  
  To take advantage of the original document hidden states $h_t^{X^d}$ and the polished document hidden states $h_t^{p}$, we combine them both into headline generation by a dynamic attention:
  \begin{align}
  \delta_{it}&=\frac{{\rm exp}(f(h_i^*, d_t))}{\sum\nolimits_j {\rm exp}(f(h_j^*, d_t))}.\\
  g_t^{*}&=\sum\nolimits_i \delta_{it}h_i^*,
  \end{align}
  where $h_i^*$ can be a polished document state $h_i^{p}$ or an original document state $h_i^{X^d}$.
  The matching function $f$ is designed as $f = h_i^* W_f d_t$ which is simple but efficient.
  When combining $g_t^{X^d}$ and $g_t^{p}$, we use an ``editing gate'' $\gamma$, which is determined by the decoder state $d_t$:
  \begin{gather}
  \gamma=\sigma(W_g d_t + b_g),\\
  h^E_t=\gamma g_t^{X^d} + (1-\gamma) g_t^{p} \label{contextvector},
  \end{gather}
  where $\sigma$ denotes the sigmoid function. 
  $h^E_t$ is further concatenated with the decoder state $d_t$ and fed into a linear layer to obtain the decoder output state $d_t^o$:
  \begin{gather}
  d_t^o=W_o[d_t;h^E_t] + b_o.
  \end{gather}
  To incorporate the guidance of the extracted style representation $s$, we combine the decoder output state with $s$ using a ``guidance gate'' $\gamma_s$:
  \begin{gather}
  \gamma_s = \sigma(W_g d_{t} + b_g),\\
  \hat{d}_t^o = \gamma_s d_t^o + (1-\gamma_s) s.
  \end{gather}
  In this way, the decoder can automatically decide the extent to which the attractive style representation is incorporated. 
  Finally, we obtain the generated word distribution $P_v$:
  \begin{gather}
  P_v={\rm softmax}(W_v \hat{d}_t^o+b_v).
  \end{gather}
  The loss is the negative log likelihood of the target word $y_t$:
  \begin{gather}
  \mathcal{L}_{seq} = -\sum_{t=1}^{n^d} \log P_v(y_t).
  \end{gather}
  
  To handle the out-of-vocabulary problem, we also equip our decoder with a pointer network~\cite{see2017get}.     
As our model is trained in an adversarial manner, we separate our model parameters into two parts: 

(1) \textit{discriminator module} contains the parameters of the discriminator in the style space constraint, which is optimized by $\mathcal{L}_D$ as:
\begin{gather}
    \mathcal{L}_D=\mathcal{L}_S^d + \mathcal{L}_C^d. \label{dd}
\end{gather}
(2) \textit{generation module} consists of all other parameters in the model, optimized by $\mathcal{L}_G$ calculated as:
\begin{gather}
\mathcal{L}_G=\mathcal{L}_{VAE} + \mathcal{L}_{Cs} + \mathcal{L}_{Cc} + \mathcal{L}_S^g + \mathcal{L}_C^g +  \mathcal{L}_{seq}.\label{L_G}
\end{gather}

  \section{Experimental Setup}
  
  
  
  \subsection{Dataset}
  \label{dataset}
  We use the public Tencent Kuaibao Chinese News dataset\footnote{\url{https://kuaibao.qq.com/}} proposed by~\citet{qin2018automatic}. 
  The dataset contains 160,922 training samples, 1,000 validation samples, and 1,378 test samples. 
	All these samples have more than 20 comments, which can be regarded as attractive cases \cite{xu2019clickbait}.
  Since we use unattractive headlines to assist in the disentanglement of content and style, we collect 1,449 headlines with less than 20 comments from the same Kuaibao website as an unattractive headline dataset.
  The vocabulary size is set to 100k for speed purposes. Lucene\footnote{\url{https://lucene.apache.org}} is employed to retrieve the prototype document-headline pair.
  Concretely, the document-headline pair in the training dataset that is most similar to the current document-headline pair is selected as prototype pair.
  The similarity of document-headline pairs is calculated by the scalar product of the TF-IDF vector~\cite{luhn1958automatic} of each document. 
  
  \subsection{Comparisons}
  \label{comparison}
  We compare our proposed method against several baselines:
(1) \textbf{Lead}~\cite{nallapati2017summarunner, see2017get} selects the first sentence of document as the summary.
(2) \textbf{Proto} directly uses the retrieved headline as the generated one.
(3) \textbf{PG}~\cite{see2017get} is a sequence-to-sequence framework with attention mechanism and pointer network.
(4)\textbf{R\textsuperscript{3}Sum} extends the seq2seq framework to jointly conduct template-aware summary generation~\cite{cao2018retrieve}.
(5) \textbf{Unified}~\cite{hsu2018unified} combines the strength of extractive and abstractive summarization.
(6) \textbf{PESG}~\cite{gao2019write} generates summarization through prototype editing.
(7) \textbf{SAGCopy}~\cite{xu2020self} is a augmented Transformer with self-attention guided copy mechanism.
(8) \textbf{GPG} generates headlines by ``editing'' pointed tokens instead of hard copying~\cite{shen2019improving}.
(9) \textbf{Sensation} generates attractive headlines using reinforcement learning method~\cite{xu2019clickbait}.
(10) \textbf{SLGen}~\cite{zhang2020structure}, encodes relational information of sentences and automatically learns the sentence graph for headline generation.

  \subsection{Evaluation Metrics}
  \textbf{ROUGE:} We evaluate models using standard full-length ROUGE F1~\cite{lin2004rouge} following previous works~\cite{gao2019write, xu2019clickbait}. 
ROUGE-1, ROUGE-2, and ROUGE-L refer to the matches of unigram, bigrams, and the longest common subsequence, respectively.
  
  \textbf{BLEU:} To evaluate our model more comprehensively, we also use the metric BLEU proposed by~\citet{papineni2002bleu} which measures word overlap between the generated text and the ground-truth.
  We adopt BLEU-1$\sim$4 and BLEU which adds a penalty for length.
  
  \textbf{Human evaluation:} 
  \citet{schluter2017limits} noted that only using the autometric to evaluate generated text can be misleading.
  Therefore, we also evaluate our model by human evaluation. 
We randomly sample 100 cases from the test set and ask three different educational-levels annotators to score the headlines generated by PESG, SAGCopy, Sensation, and our model DAHG.
  The statistical significance of observed differences is tested using a two-tailed paired t-test and is denoted using \dubbelop (or \dubbelneer) for strong (or weak) significance for $\alpha = 0.01$.

\begin{table*}[t]
    \centering
    \small
    \caption{BLEU and ROUGE scores comparison with baselines. All our ROUGE scores have a 95\% confidence interval of at most $\pm$0.22 as reported by the official ROUGE script.}
    \begin{tabular}{@{}lcccccccc@{}}
      \toprule
      & BLEU & BLEU-1 & BLEU-2 & BLEU-3 & BLEU-4 & R-1 & R-2 & R-L\\
      \midrule
      \emph{extractive summarization}\\
      Lead  & 4.82 & 13.07 & 4.89 & 3.27 & 2.58 & 19.70 & 7.98 & 17.41 \\
      Proto  & 6.62 & 22.72 & 7.14 & 4.09 & 2.89 & 22.22 & 6.90 & 19.61 \\
      \midrule
      \emph{abstractive summarization}\\
      PG \cite{see2017get} & 10.40 & 24.42 & 10.59 & 7.43 & 6.11 & 25.92 & 10.86 & 23.40 \\
      R\textsuperscript{3}Sum \cite{cao2018retrieve} & 11.22 & 26.49 & 11.51 & 7.95 & 6.53 & 27.78 & 11.81 & 25.02 \\
      Unified \cite{hsu2018unified} & 10.55 & 25.09 & 10.85 & 7.51 & 6.06 & 27.94 & 11.68 & 25.37 \\
      PESG \cite{gao2019write} & 11.08 & 26.62 & 11.43 & 7.84 & 6.33 & 28.21 & 11.87 & 25.56   \\
      SAGCopy \cite{xu2020self} & 11.21 & 27.65 & 11.33 & 7.75 & 6.50 & 28.71 & 11.27 & 25.74   \\
      \midrule
      \emph{headline generation}\\
GPG \cite{shen2019improving}  & 10.51 & 24.48 & 11.00 & 7.60 & 5.96 & 26.12 & 11.61 & 23.84 \\
Sensation \cite{xu2019clickbait} & 10.70 & 26.18 & 10.89 & 7.54 & 6.10 & 26.28 & 10.84 & 23.94\\
SLGen \cite{zhang2020structure} & 11.48 & 26.50 & 11.76 & 8.28 & 6.74 & 27.33 & 11.83 & 25.00\\
      \midrule
DAHG &  \textbf{12.74} &  \textbf{28.93} &  \textbf{13.34} &  \textbf{9.27} &  \textbf{7.38} &  \textbf{29.73} &\textbf{13.55} &  \textbf{27.03} \\
\quad+BERT & 11.43 & 27.89 & 12.06 & 7.93 & 6.74 & 28.61 & 12.27 & 24.08\\

      \bottomrule
    \end{tabular}
    \label{tab:comp_baslines}
  \end{table*}

  \subsection{Implementation Details}
  
  Our experiments are implemented in Tensorflow~\cite{abadi2016tensorflow} on an NVIDIA GTX 1080 Ti GPU. 
  Experiments are performed with a batch size of 64.
  We pad or cut the input document to 400 words and the prototype headline to 30 words.
  The maximum decode step is set to 30, and the minimum to 10 words.
  We initialize all of the parameters in the model using a Gaussian distribution. 
  We choose Adam optimizer for training, and use dropout in the VAE encoder with keep probability as 0.8. 
  For testing, we use beam search with size 4 to generate a better headline.
  It is well known that a straightforward VAE with RNN decoder always fails to encode meaningful information due to the vanishing latent variable problem \cite{bowman2015generating}.
  Hence, we use the BOW loss along with KL annealing of 10,000 batches to achieve better performance.
   Readers can refer to \cite{zhao2017learning} for more details.

  \section{Experimental Result}
  \subsection{Overall Performance}
  
We compare our model with the baselines in Table~\ref{tab:comp_baslines}.
Firstly, Proto outperforms Lead but underperforms other baselines, indicating that prototype information is helpful for our task, but directly taking it as input only leads to a small improvement.
  Secondly, abstractive methods outperform all extractive methods, demonstrating that Kuaibao is a dataset suitable for abstractive summarization.
Finally, our model outperforms PESG by 14.99\%, 5.39\%, 14.15\%, 5.75\%, and outperforms SAGCopy by 13.65\%, 3.55\%, 20.23\%, 5.01\% in terms of BLEU, ROUGE-1, ROUGE-2, and ROUGE-L, respectively, which proves the superiority of our model.
Besides, it can also be seen that with BERT augmented as the encoder of our model, the results are slightly below DAHG, which demonstrates that simply superimposing the pre-training module on our model does not help improve the effect.

  For the human evaluation, we ask annotators to rate headlines generated by PESG, SAGCopy, Sensation, and our model DAHG according to fluency, consistency, and attractiveness.
  The rating score ranges from 1 to 3, with 3 being the best. 
  Table~\ref{tab:comp_human_baslines} lists the average scores of each model, showing that DAHG outperforms other baseline models among all metrics.
  Besides, to directly compare the attractiveness of each model, we ask annotators to select the headline they are most likely to click among the four candidate headlines.
  The click-rate is also listed in Table~\ref{tab:comp_human_baslines}, where our model wins 22\% click among all baselines.
This direct comparison shows that our headlines are more eye-catching and stand out above the rest.
Note that, the generated headlines of DAHG are not clickbait, since they are generally faithful to the content of the documents as the consistency score shows.
  The kappa statistics are 0.49, 0.54, and 0.47 for fluency, consistency, and attractiveness, respectively, which indicates the moderate agreement between annotators.
  To verify the significance of these results, we also conduct the paired student t-test between our model and Sensation (the row with shaded background).
  We obtain a p-value of $5 \times 9^{-9}$, $3 \times 10^{-7}$, and $5 \times 10^{-6}$ for fluency, consistency, and attractiveness. 
  
  \begin{table}[t]
    \centering
    \small
    \caption{Fluency(Flu), consistency(Con), attractiveness(Attr) and click-rate(Clr) comparison by human evaluation.}
    \begin{tabular}{@{}lcccc@{}}
      \toprule
      & Flu & Con & Attr & Clr \\
      \midrule
      PESG & 2.07 &2.14  & 1.70 &0.11  \\
      SAGCopy & 2.21  &  2.08& 1.93 & 0.13\\
      \cbkgrnd Sensation &\cbkgrnd 2.02  &\cbkgrnd 1.97  &\cbkgrnd 2.19  &\cbkgrnd 0.27  \\
      DAHG & \textbf{2.59}\dubbelop &\textbf{2.51}\dubbelop  & \textbf{2.47}\dubbelop &\textbf{0.49}  \\
      \bottomrule
    \end{tabular}
    \label{tab:comp_human_baslines}
  \end{table}

  \subsection{Ablation Study}
In order to verify the effect of each module in DAHG, we conduct ablation tests in Table~\ref{tab:ablation_study}.
  We first verify the effectiveness of separating style and content in DAHG-D, where we directly use the prototype headline to polish the input document.
  DAHG-C omits the process of polishing by attractive content representation, and DAHG-S does not use attractive style to guide the generation process.
  All ablation models perform worse than DAHG in terms of all metrics, demonstrating the preeminence of DAHG.
  Specifically, the polishing process contributes most to DAHG, and separating style and content is also important in achieving high performance.
  

  \subsection{Analysis of Disentanglement}
  
\begin{figure} 
    \centering 
    \subfigure[]{ 
      \includegraphics[width=0.45\linewidth]{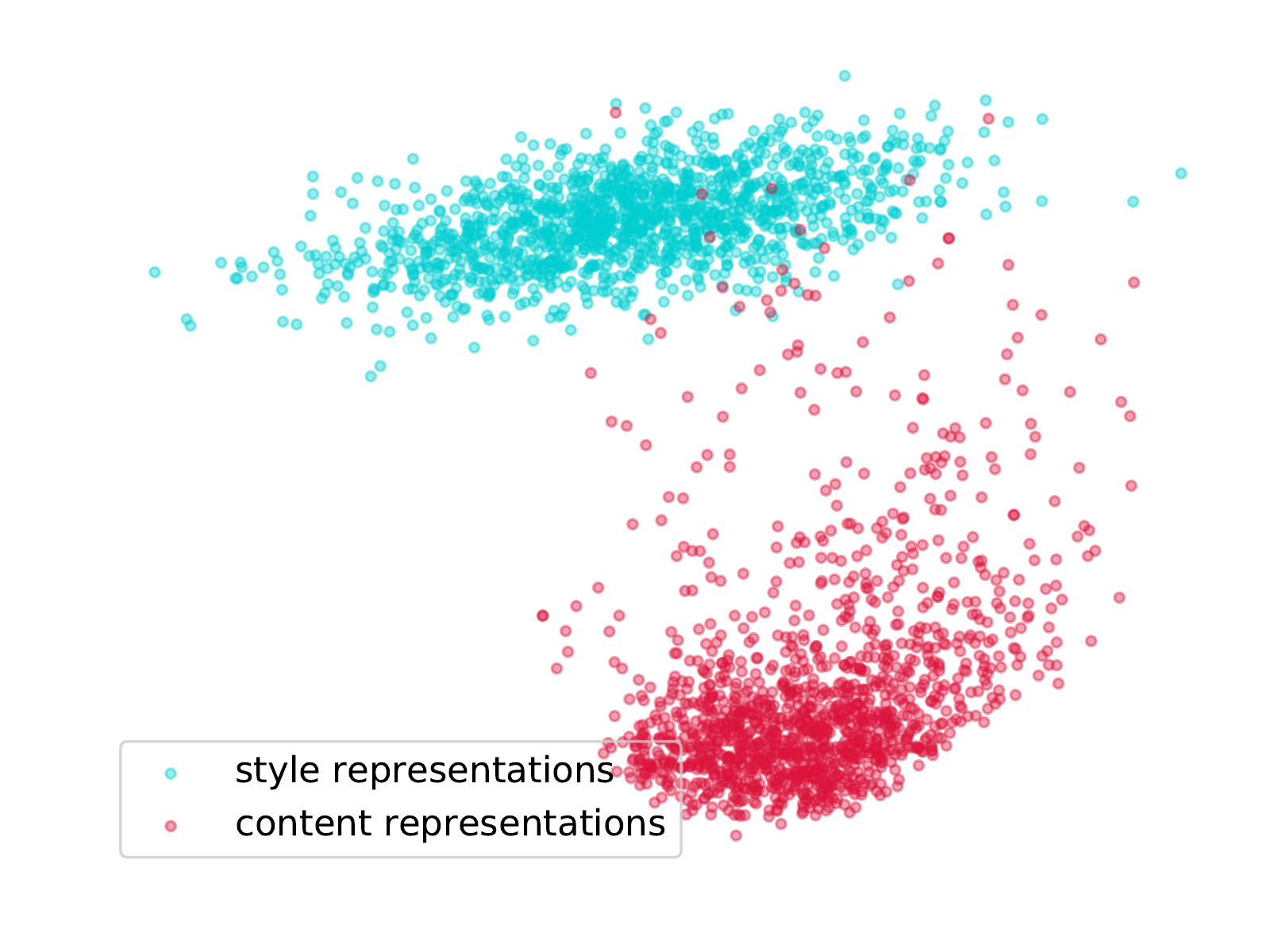}
    }
    \subfigure[]{ 
      \includegraphics[width=0.45\linewidth]{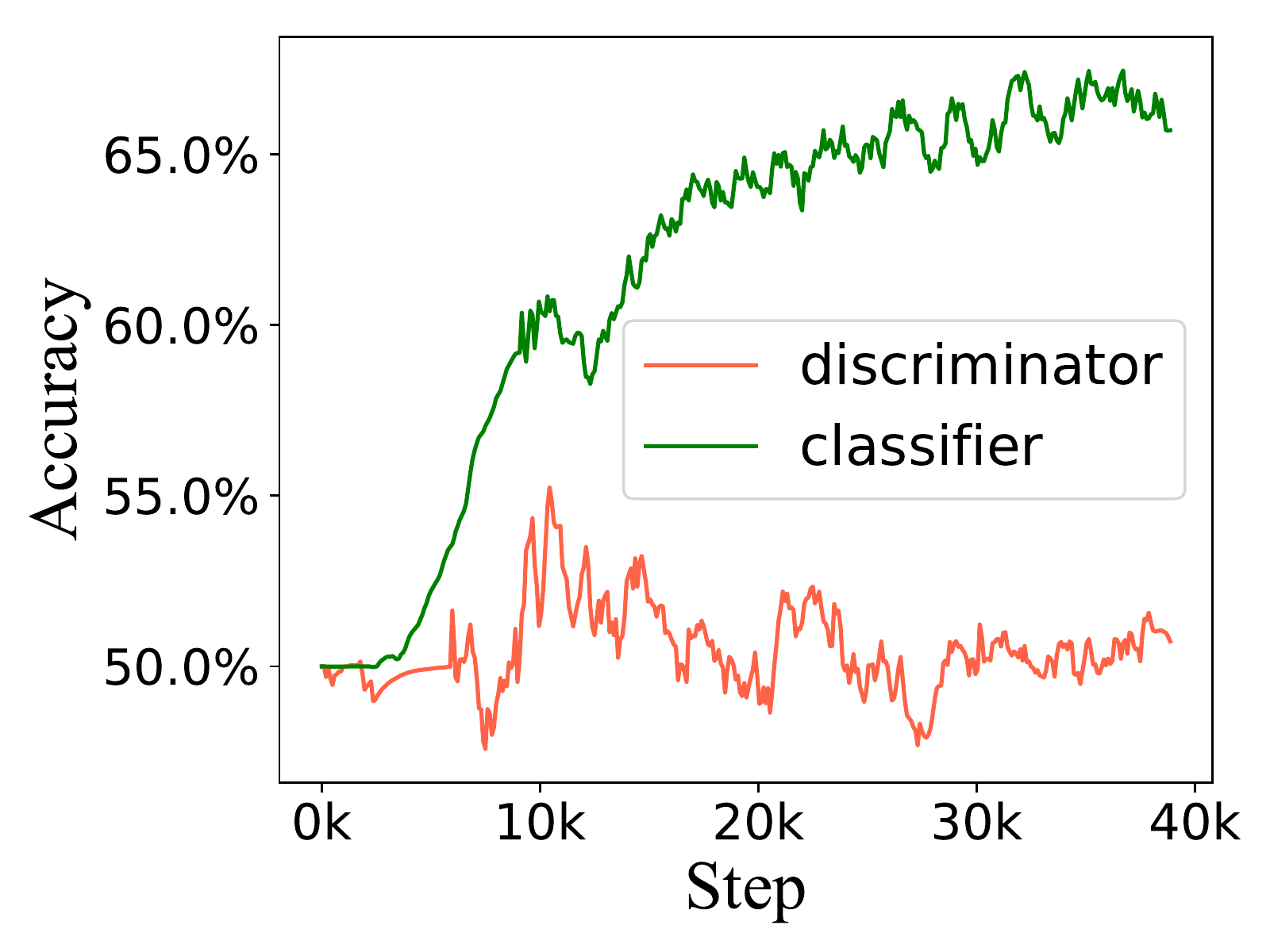}
    }
    \caption{(a) Visualization result for disentanglement. (b) Accuracy curve of classifier and discriminator in style space constraint.} 
    \label{fig:subfig}
  \end{figure}

The vectors in the style and content spaces on test data are visualized in Figure~\ref{fig:subfig}(a). For visualization purpose, we reduce the dimension of the latent vector with t-SNE \cite{maaten2008visualizing}. It can be observed that points representing same space located nearby in the 2D space while different ones. This suggests that the disentanglement module can disentangle the prototype headline into style and content space.
 \begin{table}[t]
    \centering
    \small
    \caption{ROUGE scores of different ablation models of DAHG.}
    \begin{tabular}{@{}lccc@{}}
      \toprule
      & R-1 & R-2 & R-L \\
      \midrule
      All & 29.73 & 13.55 & 27.03\\
      \midrule
      DAHG-D & 27.69 & 12.01 & 25.32\\ 
      DAHG-C & 27.60 & 11.97 & 24.93\\
      DAHG-S & 27.82 & 12.43 & 25.32\\
      \bottomrule
    \end{tabular}
    \label{tab:ablation_study}
  \end{table}
  
Besides, the accuracy curves of classifier and discriminator in style space constraint are shown in Figure~\ref{fig:subfig}(b). When the training begins, the accuracy of the discriminator fluctuates from time to time, which verifies the adversarial training. After several steps, the training converges, and the accuracy of the discriminator stays around 50\%, which means that the content information is successfully excluded from style space.
The discriminator cannot perform better than a random guess.
As for the accuracy curve of the classifier, the accuracy score of choosing the style label is on a steady upward curve and finally reaches convergence along the training progresses, which proves that the style information is contained in the style space.
The accuracy curves in content space constraint are similar, and are omitted due to the limited space.

  \begin{table}[t]
    \centering
    \small
    \caption{Case study to verify the influence of sampling on style and content space.
      The text in blue denotes attractive information related to the prototype headline, and text in red denotes information that is related to the prototype but not proper for a headline.
    }
    \begin{tabular}{c|l}
      \toprule
      \multicolumn{2}{p{8cm}}{    \emph{\textbf{Document:}}
        I believe everyone is familiar with Xu Xiaodong in the picture above. 
        A fight between modern boxing and traditional Taiji made him become a social focus. 
        Today, another containment incident happened, which made Xu a little nervous.
        At 16:25 this afternoon, Xu was contained by seven Taiji students during the live broadcast.
        \hlc[cyan!50]{When talking about the situation at that time, Xu said: "Seven students won't let me go. I'm afraid."}
        When asked what he thought of when he was surrounded, Xu said, ``I worried about what will happen to my assistant? \hlc[pink]{My assistant is a girl, I cannot fight back}. So I call the police.''. 
        Although Xu is sometimes reckless, his original intention is admirable.
        He did not try to fight against the martial, instead, he only meant to fight against the cheaters in the traditional martial arts. 
      } \\ \hline 
      
      \multicolumn{2}{p{8cm}}{
        \emph{\textbf{Reference headline:}}\textcolor{black}{Emergencies! Xu Xiaodong is contained by seven Taiji disciples.}
      }\\ \hline
      \multicolumn{2}{p{8cm}}{
        \textbf{Case 1:}
      }\\
       \multicolumn{2}{p{8cm}}{
        \emph{\textbf{Prototype 1:}}
        Xu Xiaodong behaves beyond the bottom line and has been despised by the whole traditional martial arts.    
      }\\
      \multicolumn{2}{p{8cm}}{
        \emph{\textbf{DAHG 1:}}
        \hlc[cyan!50]{Xu Xiaodong is contained by Taiji disciples}  and calls the police! Saying that he might have accidents.
      }\\ \hline
       \multicolumn{2}{p{8cm}}{
        \textbf{Case 2:}
      }\\
       \multicolumn{2}{p{8cm}}{
        \emph{\textbf{Prototype 2:}}
        What are the most beautiful girls like in a boy's eyes?
      }\\
      \multicolumn{2}{p{8cm}}{
        \emph{\textbf{DAHG 2:}}
        Xu Xiaodong: \hlc[pink]{My assistant is a girl. I cannot fightback.}
      }\\ \hline

    \multicolumn{2}{p{8cm}}{
        \emph{\textbf{PESG:}}
        Xu Xiaodong: a modern fight vs. Traditional Taiji makes him the focus of the society.}\\ \hline
    \multicolumn{2}{p{8cm}}{
        \emph{\textbf{SAGCopy:}}
        Xu Xiaodong: I am a fighting madman, and the person I fear most is myself.}\\ \hline
      \multicolumn{2}{p{8cm}}{
        \emph{\textbf{Sensation:}}
        Xu Xiaodong fights against all Taiji students in the Wulin. My favorite is mine.}\\
      \bottomrule
    \end{tabular}
    \label{tab:sample-case}
  \end{table}
  

  \subsection{Analysis of Sampling Effect}
  We finally investigate the influence of sampling different latent variables in the attractive style and content space.
  As shown in Table~\ref{tab:sample-case}, for a single document, we choose one similar prototype headline and one randomly-selected prototype headline as the input of DAHG, and examine the quality of the generated headline.
  Different prototype headlines lead to different sampling results in the style and content space, and correspondingly, the more proper the prototype is, the better the sampling result and the headline are.
  Case 1 in Table~\ref{tab:sample-case} is a good sampling case, where prototype 1 is selected as the most similar one.
  We can see that the style and content of the generated headline match the document well, emphasizing on ``be contained by students''.
  As for case 2, the prototype focuses on gender information, hence leads to a bad sampling result, emphasizing ``assistant is a girl'', which does not cover the necessary information of a proper headline.

  We also compare our model with several baselines in Table~\ref{tab:sample-case}.
  Most baselines can generate fluent headlines in this case.
  However, they miss the attractive style and can include unattractive content.
  The headline generated by PESG is a plain statement.
  Sensation generates more attractive headline, but is not faithful to the document, and includes unnecessary content.
  While for our model in case 1, DAHG captures the keywords ``contained'' and ``police'' that not only cover the events in the document but also draw mass attention.  
  The full version with the original Chinese version of the case study can be found in Appendix~\ref{appendix}.
  Representative attractive headlines generated by DAHG can be found in Appendix~\ref{appendix2}.

  \section{Conclusion}
  In this paper, we propose a Disentanglement-based Attractive Headline Generator (DAHG) to generate an attractive headline. 
  Our model is built on the fact that the attractiveness of the headline comes from both style and content aspects.
  Given the prototype document-headline pair, DAHG disentangles the attractive content and style space from the prototype attractive headline. 
  The headline generator generates attractive headlines under the guidance of both. 
  Our model achieves state-of-the-art results in terms of ROUGE scores and human evaluations by a large margin.
  In near future, we aim to bring the model online.

 \section*{Ethical Impact Statement}
 The motivation of our paper is to generate attractive headline for news, which is useful for both readers and writers. On one hand, generating headlines that can trigger high click-rate is especially important for different avenues and forms of media to compete for user's limited attention. On the other hand, only with the help of a good headline, the outstanding article can be discovered by readers.
 
The generated headlines sometimes can sometimes be cilckbaits, which cause users to fall into information garbage. However, theoretically, all headline generation modules face the same problem. In the model design, we maintain the consistency of the headline's and the document's content information as much as possible, so as to ensure that the generated headline is faithful to the content of the document to a large extent. In the future, we will continue to explore how to enhance the consistency of content in the headline generation.


\section*{Acknowledgments}
We would like to thank the anonymous reviewers for their constructive comments.
This work was supported by the National Key Research and Development Program of China (No. 2020YFB1406702), the National Science Foundation of China (NSFC No. 61876196 and No. 61672058), Beijing Outstanding Young Scientist Program (No. BJJWZYJH012019100020098). Rui Yan is supported as a young fellow at Beijing Academy of Artificial Intelligence (BAAI).


  \clearpage
  \appendix
\section{SRU Cell}
\label{sec:sru}

\begin{figure}[H]
    \centering
    \includegraphics[scale=0.6]{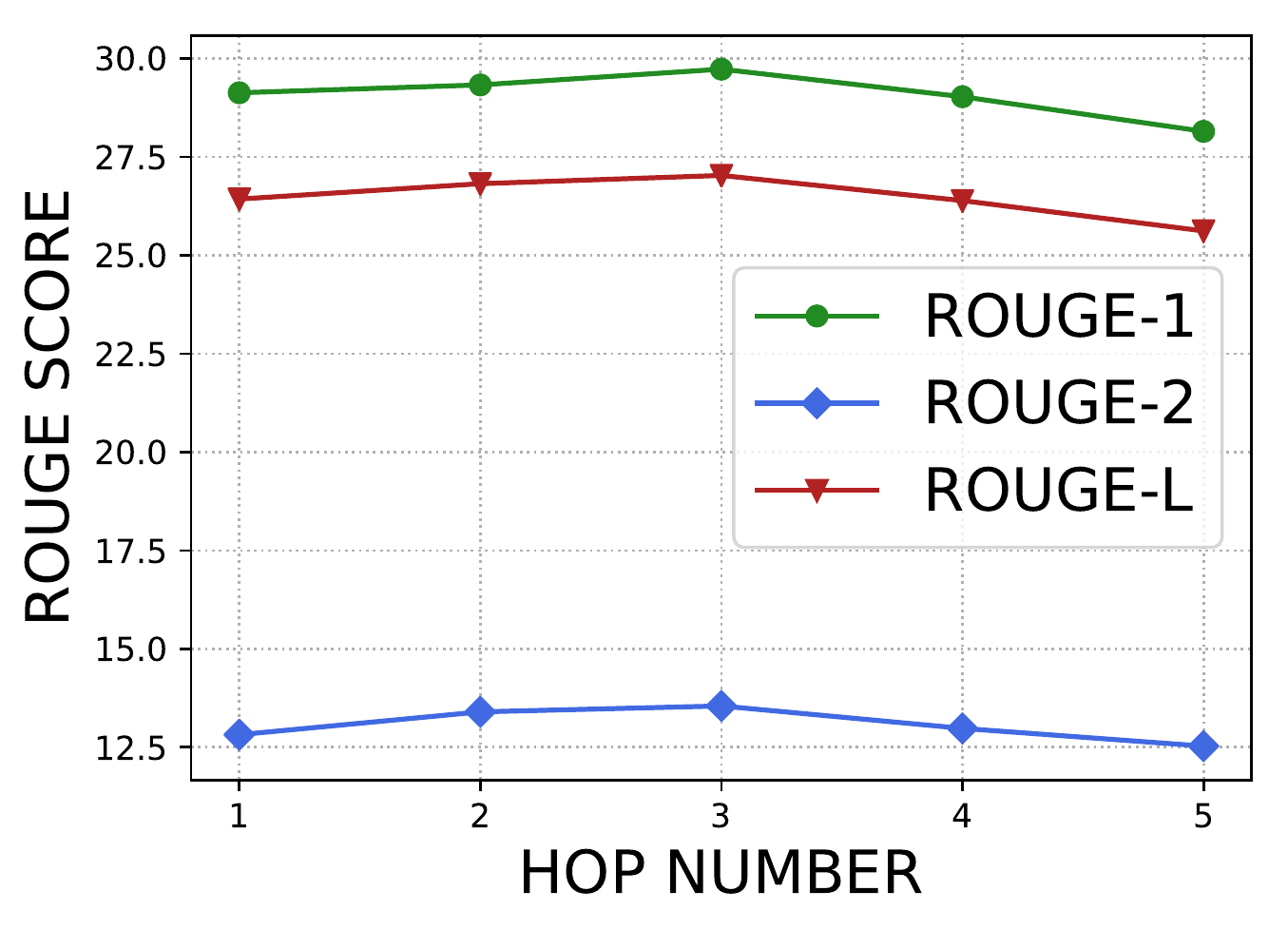}
    \caption{
      The effectiveness of different hop number.
    }
    \label{fig:fgru-hop}
  \end{figure}

Gated recurrent unit (GRU)~\cite{Cho2014LearningPR}, which combines an update gate in RNN, is a gating mechanism in recurrent neural networks. Here first is the details of the original GRU.
\begin{align}
u_{t} &= \sigma(W_{u}^{1}x_{t}+W_{u}^{2}h_{t-1}+b_{u}),  
\label{gru-gated} \\
r_{t} &= \sigma(W_r^1x_{t}+W_r^2h_{t-1}+b_r),  \\
\hat{h_{t}} &= \text{tanh}(W_h^1x_{t}+r_{t}\circ W_h^2h_{t-1}+b_h), \\
h_{t} &= u_{t}\circ\hat{h_{t}}+(1-u_{t})\circ h_{t-1}, \label{gru-hi}
\end{align}
where $\sigma$ is the sigmoid activation function, $W_u^1, W_r^1, W_h^1 \in \mathbb{R}^{n_{H}\times n_{I}},W_u^2, W_r^2, W_h^2\in \mathbb{R}^{n_{H}\times n_{H}}, n_{H}$ is the hidden size, and $n_{I}$ is the size of input $x_{t}$.
In the original version of GRU, the update gate $u_t$ in Equation~\ref{gru-gated} decides how much of the hidden state should be updated and how much should be retained.
In our highlight polish module, we want to utilize the polished facts $h_t^k$ at the $k$-th hop as $h_t^p$ to decide which facts are attractive. To achieve this, we modify the calculation of $u_t$ using the newly computed update gate $g_t$:
\begin{align}
e_{t} &= [h_{t}^{k-1} \circ c; h_{t}^{k-1}; c], \\
z_{t} &= W^{(2)}\tanh(W^{(1)}e_{t}+b^{(1)})+b^{(2)}, \\
g_{t} &= \frac{\exp(z_{t})}{\sum^{n_{s}}_{j=1} \exp(z_{j})},
\end{align}
where $W^{(2)}, W^{(1)}, b^{(1)}, b^{(2)}$ are all trainable parameters and $k$ is the hop number in the multi-hop situation which is a hyper-parameter manually set.
The effectiveness of different hop number is verified in the experimental results shown in Figure~\ref{fig:fgru-hop}.
Equation~\ref{gru-hi} now becomes:
\begin{align}
h_{t} &= g_{t} \circ \hat{{h_{t}}}+(1-g_{t}) \circ h_{t-1 }.
\end{align}
The name ``SRU'' in ~\S~\ref{sec:polish} corresponds to this modified version of GRU cell.

  \section{Case Study}
  \label{appendix}
  \begin{CJK*}{UTF8}{gkai}
    \begin{table}[H]
      \centering
      \small
      \caption{Case study to verify the influence of sampling on style and content space.
        The text in blue denotes attractive information related to the prototype headline, and text in red denotes information that is related to the prototype but not proper for a headline.
      }
      \begin{tabular}{l|l}
        \toprule
        \multicolumn{2}{p{8cm}}{      
          \emph{\textbf{Document:}}
          I believe everyone will be familiar with Xu Xiaodong in the picture above. 
          A fight between modern boxing and traditional Taiji made him become a social focus! 
          Today, another containment incident happened, which made Xu a little nervous.
          At 16:25 this afternoon, Xu was surrounded by seven Taiji students during the live broadcast.
          \textcolor{blue}{When talking about the situation at that time, Xu said: ``Seven students won't let me go. I'm afraid.''}
          When asked what he thought of when he was surrounded, Xu said, ``what will happen to my assistant? \textcolor{red}{My assistant is a girl, I cannot fight back}. \textcolor{black}{So I call the police}''. 
          Although Xu is reckless in doing things, his original intention is admirable.
          He did not try to fight against the traditional martial arts. 
          He only meant to fight against the cheaters in the traditional martial arts. 
          相信大家对上图这位徐晓冬先生不会陌生，一场现代搏击vs传统太极让他成为社会焦点人物。今天又发生一件围堵事件，使得徐晓冬有些惶恐不安。今天下午16：25，徐晓冬
          在做直播的时候遭到7名陈氏太极学生的围堵。\textcolor{blue}{谈到当时的情景时徐晓冬说：“7个学生不让我走。”}当被问到，自己被围住时想到是是什么，徐晓冬说：“我
          的助理怎么办，\textcolor{red}{我的助理是个女孩，我不能上手打人}。\textcolor{black}{所以我选择了报警 }。”虽然徐晓冬做事情有些鲁莽，但出发点还是值得赞赏的。他的本意也不是说要打压传统武术，他的意思打假传统武术中的骗子。
        } \\ \hline 
        
        \multicolumn{2}{p{8cm}}{
          \emph{\textbf{Reference headline:}}
          
          Emergencies! Xu Xiaodong is surrounded by 7 Taiji disciples.(突发事件！徐晓冬遭7名陈氏太极学生围堵)
        }\\ \hline
        \multicolumn{2}{p{8cm}}{
          \textbf{Case 1:}
        }\\
        \multicolumn{2}{p{8cm}}{
          \emph{\textbf{Prototype 1:}}
          
          Xu Xiaodong behaves beyond the bottom line and has been despised by the whole traditional martial arts. (徐晓东 的 零 底线 行为 已经 让 整个 传统 武学 所 唾弃)    
          }\\
        \multicolumn{2}{p{8cm}}{
          \emph{\textbf{Generated headline 1:}}
          
          \textcolor{blue}{Xu Xiaodong is surrounded by Taiji disciples} and calls the police! Saying that he might have
          accidants.(\textcolor{blue}{徐晓冬遭太极弟子围堵}并报警！坦言自己或出意外
        } \\ \hline
        \multicolumn{2}{p{8cm}}{
          \textbf{Case 2:}
        }\\
        \multicolumn{2}{p{8cm}}{
          \emph{\textbf{Prototype 2:}}
          
          What are the most beautiful girls like in a boy's eyes? (男生眼中女生什么时候最美？)
          }\\
        \multicolumn{2}{p{8cm}}{
          \emph{\textbf{Generated headline 2:}}
          
          Xu Xiaodong: \textcolor{red}{ My assistant is a girl. I cannot fightback.} (徐晓冬：\textcolor{red}{ 我的助理是个女孩，我不能上手打人})
        } \\ \hline
         \multicolumn{2}{p{8cm}}{
          \emph{\textbf{PESG:}}
          Xu Xiaodong: a modern fight vs. Traditional Taiji makes him the focus of the society.(徐晓冬 ： 一场 现代 搏击 vs 传统 太极 让 他 成为 社会 焦点人物)}\\ \hline
        \multicolumn{2}{p{6.5cm}}{
          \emph{\textbf{SAGCopy:}}
          Xu Xiaodong: I am a fighting madman, and the person I fear most is myself. (徐晓冬 ： 我 是 格斗 狂人 ， 我 最 害怕 的 是 我 )}\\ \hline
        
       \multicolumn{2}{p{8cm}}{
          \emph{\textbf{Sensation:}}
          Xu Xiaodong fights against all Taiji students in the Wulin. My favorite is mine. (徐晓冬 单挑 整个 武林 太极 学生 ， 我 最 爱 的 是 我 的)}\\
        \bottomrule
      \end{tabular}
      \label{tab:full-case}
    \end{table}
    
  \end{CJK*}
  
  \section{Representative Generated Results}
  \label{appendix2}
  \begin{table}[H]
    \centering
    \small
    \caption{Examples of the generated attractive headline by DAHG and sensation.
    }
    \begin{tabular}{l|l}
      \toprule
      Case 1 & \multicolumn{1}{p{6cm}}{  
        \emph{\textbf{DAHG:}}
        
        "accidents continue in a small town", two homicides have taken place!   
        
        \emph{\textbf{Sensation:}}
        
        A major traffic accident occurred at the intersection of group 6, Daliang village, Qianwei Town, Lantian County}
      \\ \hline
      Case 2 & \multicolumn{1}{p{6cm}}{  
        \emph{\textbf{DAHG:}}
        
        Biography of Chu Qiao: Zhao Liying's fighting with wolves, expertly posed for photos!    
        
        \emph{\textbf{Sensation:}}
        
        Zhao Liying fights with the wolf. Zhao Liying, do you know you?}
      \\ \hline
      Case 3 & \multicolumn{1}{p{6cm}}{  
        \emph{\textbf{DAHG:}}
        
        "Catch demons 2" was withdrawn, divorce but still pretend to be lovers for money? 
        
        \emph{\textbf{Sensation:}}
        
        Bai Baihe is stopped again, Bai Baihe is derailed again, and the director is already crying}
      \\ \hline
      Case 4 & \multicolumn{1}{p{6cm}}{  
        \emph{\textbf{DAHG:}}
        
        Golden Retriever: I tore my home and chewed my slippers. What can you do for me?   
        
        \emph{\textbf{Sensation:}}
        
        The Golden Retriever is proud, the owner hasn't found socks for a long time}
      \\ \hline
      Case 5 & \multicolumn{1}{p{6cm}}{  
        \emph{\textbf{DAHG:}}
        
        Nanning customs launched a joint action against automobile smuggling, worth about 1.9 billion yuan  
        
        \emph{\textbf{Sensation:}}
        
        Joint smuggling operation, the fake license plate of a smuggled vehicle, a smuggled vehicle under investigation}
      \\
      \bottomrule
    \end{tabular}
    \label{tab:show-case}
  \end{table}

\end{document}